\documentclass[runningheads]{llncs}
\usepackage{graphicx}
\usepackage{amsmath}

\usepackage{bm}       
\usepackage{amssymb}  

\usepackage{graphicx}
\newcommand{\CosSim}{\operatorname{CosSim}}

\begin{document}
\title{OptiPrune: Boosting Prompt-Image Consistency with Attention-Guided Noise and Dynamic Token Selection}

\titlerunning{Abbreviated paper title}

\author{Ziji Lu\inst{1}}
\authorrunning{F. Author et al.}
%
\institute{School of Computational Science \& Education Software, Guangzhou University, 510006, Guangzhou, PR China\\
\email{zijilu2025@e.gzhu.edu.cn}}
\maketitle

\begin{abstract}

Text-to-image diffusion models often struggle to achieve accurate semantic alignment between generated images and text prompts while maintaining efficiency for deployment on resource-constrained hardware. Existing approaches either incur substantial computational overhead through noise optimization or compromise semantic fidelity by aggressively pruning tokens. In this work, we propose OptiPrune, a unified framework that combines distribution-aware initial noise optimization with similarity-based token pruning to address both challenges simultaneously. Specifically, (1) we introduce a distribution-aware noise optimization module guided by attention scores to steer the initial latent noise toward semantically meaningful regions, mitigating issues such as subject neglect and feature entanglement; (2) we design a hardware-efficient token pruning strategy that selects representative base tokens via patch-wise similarity, injects randomness to enhance generalization, and recovers pruned tokens using maximum similarity copying before attention operations. Our method preserves the Gaussian prior during noise optimization and enables efficient inference without sacrificing alignment quality. Experiments on benchmark datasets, including Animal-Animal, demonstrate that OptiPrune achieves state-of-the-art prompt-image consistency with significantly reduced computational cost.

\end{abstract}

\section{Introduction}
Text-to-image diffusion models~\cite{rombach2022high,blattmann2023stable} have shown remarkable success in generating high-quality images from natural language descriptions. However, two major challenges remain, particularly in resource-constrained scenarios: (1) semantic misalignment, where generated content fails to accurately reflect the input prompt; and (2) computational inefficiency, due to the heavy denoising UNets and iterative sampling inherent in diffusion processes. These limitations hinder real-time deployment on edge devices and reduce practical usability. Addressing both issues simultaneously is thus critical for advancing efficient and reliable generative systems.

One emerging line of research tackles semantic misalignment by optimizing the initial noise fed into the diffusion process. Methods such as InitNO~\cite{guo2024initno}, ReNO~\cite{eyring2024reno}, Golden Seed Selection~\cite{xu2024good}, and PNS~\cite{shen2025imagharmony} modify or select initial latent codes that steer the model toward more semantically accurate outputs. These techniques rely on auxiliary signals, such as attention heatmaps, vision-language scores, or human preference feedback, and typically involve gradient-based optimization. While effective, their iterative refinement introduces considerable computational overhead, limiting their deployment efficiency.

In parallel, another research direction focuses on accelerating inference by reducing token redundancy. Two main strategies dominate: token merging, which aggregates similar tokens~\cite{bolya2023token,kim2024token}, and token pruning, which removes unimportant ones~\cite{wang2024attention,zhang2025training}. Though these methods significantly reduce computation, they are primarily designed for efficiency and remain largely decoupled from semantic alignment goals, often degrading generation quality when applied naively.

To bridge this gap, we propose OptiPrune—a unified and training-free framework that combines distribution-aware initial noise optimization with similarity-based token pruning to jointly improve semantic fidelity and inference efficiency. Specifically, OptiPrune consists of two synergistic components: (1) a lightweight attention-guided noise optimization module that identifies and prunes few relevant tokens before score computation, thereby reducing optimization overhead; and (2) a token pruning mechanism that selectively retains informative tokens based on similarity scores and injects controlled randomness for better generalization, followed by token recovery before denoising. This co-design enables semantic-aware acceleration throughout both the initialization and generation stages.
Comprehensive experiments on benchmarks such as Animal-Animal demonstrate that OptiPrune achieves state-of-the-art prompt-image consistency.

In summary, our contributions are as follows:  
\begin{itemize}
\item We propose a novel training-free framework that integrates initial noise optimization with similarity-based token pruning, addressing semantic misalignment and inference inefficiency in a unified design.
\item We introduce a hybrid pipeline that embeds pruning into both noise refinement and generation stages, dynamically suppressing redundant tokens while preserving semantic fidelity.
\item Our method is entirely data-agnostic, requires no fine-tuning, and generalizes well across models, datasets, and schedulers, offering a practical solution for real-time deployment.
\end{itemize}

\section{Related Work} \label{sec:rw}
\subsection{Stable Diffusion and Latent Models}
Diffusion models~\cite{ho2020denoising,song2020denoising} have emerged as a foundational approach for high-fidelity image synthesis, gradually transforming noise into visual data through iterative denoising. Latent diffusion models (LDMs)\cite{rombach2022high,vahdat2021score} reduce computational costs by operating in compressed latent spaces, enabling high-resolution image generation with tractable resources. Despite advances in quality and flexibility, diffusion-based methods remain computationally expensive due to repeated attention operations and large UNet architectures. Recent work such as IMAGPose\cite{shen2024imagpose} and IMAGGarment~\cite{shen2025imaggarment} highlights how generation quality depends heavily on accurate layout, appearance control, and prompt interpretation—all of which require significant inference-time resources. As a result, practical deployment on edge devices remains a critical bottleneck.

\subsection{Initial Noise Optimization}
Initial noise optimization aims to guide diffusion trajectories by modifying latent initialization, thereby enhancing semantic alignment with input prompts. InitNO~\cite{guo2024initno} introduces response/conflict-based latent partitioning using attention cues, while ReNO~\cite{eyring2024reno} utilizes preference learning and gradient-based noise updates. Golden Seed Selection~\cite{xu2024good} demonstrates that prompt-consistent outputs can be obtained by classifying promising seeds. IMAGHarmony~\cite{shen2025imagharmony} extends this idea to controllable editing, filtering initial noise with vision-language similarity for layout-consistent multi-object generation. Similarly, Long-TalkingFace~\cite{shen2025long} applies motion-aware initial noise tuning in conditional diffusion for video synthesis. While these methods enhance semantic controllability, they often introduce heavy optimization loops or ranking modules, significantly increasing runtime and limiting scalability.

\subsection{Token Reduction without Training}
Token reduction approaches accelerate inference by dynamically reducing the number of tokens processed during attention, often without requiring model retraining. Token merging strategies, such as ToMe~\cite{bolya2022token}, fuse similar tokens to reduce computation while preserving global structure. This has been extended in vision-language tasks by methods like Chat-UniVi~\cite{jin2024chat} and MovieChat~\cite{song2024moviechat}, which employ similarity-based merging across images and videos. In image generation, ToMeSD~\cite{bolya2023token} applies token merging before attention blocks, while ToFu~\cite{kim2024token} performs merge selection adaptively across layers. IMAGDressing~\cite{shen2025imagdressing} shows that efficient virtual try-on also benefits from spatially-aware feature aggregation in generation tasks.

Token pruning techniques, by contrast, remove redundant tokens based on learned or heuristic importance scores. Methods like AT-EDM~\cite{wang2024attention} introduce group-wise pruning and recovery to preserve fidelity, while SiTo~\cite{zhang2025training} prunes with cosine-based similarity for lightweight diffusion. Boosting-Story~\cite{shen2025boosting} explores pruning for multi-turn generation in story visualization, combining context preservation with computation reduction. Though effective for acceleration, such approaches often compromise alignment fidelity and do not adapt well to prompt-dependent content diversity.

\section{Proposed Method}\label{sec:method}

\begin{figure}[htbp]
    \centering
    \includegraphics[width=1.0\textwidth]{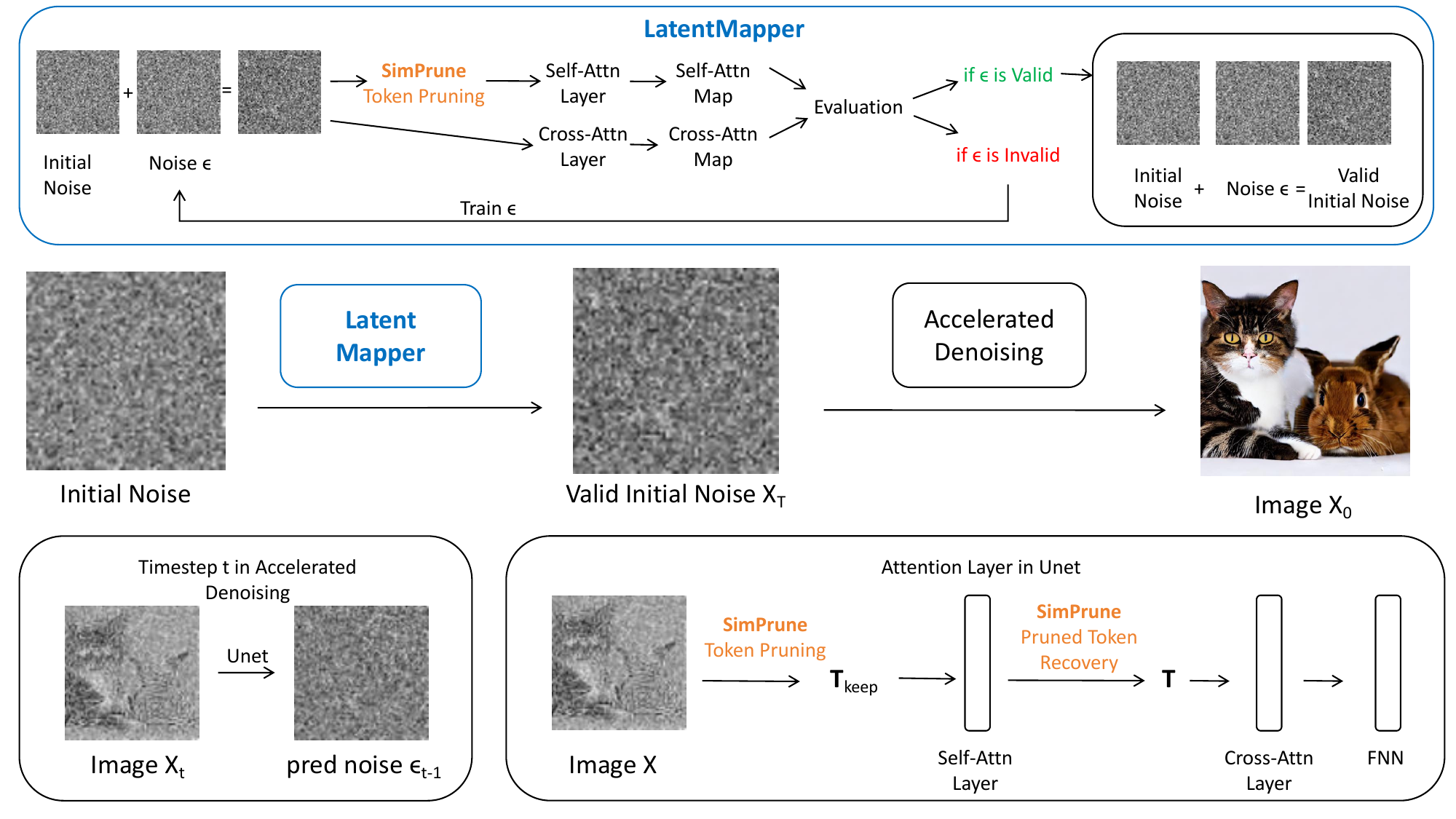}
    \caption{Illustration of our joint framework integrating LatentMapper and SimPrune. (a) LatentMapper partitions latent space using cross-attention and self-attention map to optimize initial noise toward valid regions. (b) SimPrune prunes redundant tokens in self-attention layers through spatially distributed base token selection and similarity-based recovery. The optimized noise and pruned token set are jointly leveraged for efficient, high-fidelity generation.}
    \label{fig1}
\end{figure}

\subsection{Stable Diffusion Model}

This work builds upon the Stable Diffusion (SD) model, a state-of-the-art latent diffusion model operating in the compressed latent space of a pretrained autoencoder. The autoencoder consists of an encoder $\mathcal{E}(\cdot)$ that maps an input image $\mathbf{T} \in \mathcal{X}$ to a latent representation $\mathbf{z} = \mathcal{E}(\mathbf{T})$, and a decoder $\mathcal{D}(\cdot)$ that reconstructs the image such that $\mathcal{D}(\mathcal{E}(\mathbf{T})) \approx \mathbf{T}$. 

Within this latent space, a Denoising Diffusion Probabilistic Model (DDPM) performs the generative process. During conditional generation, the model integrates textual guidance through embeddings $\mathbf{c} = f_{\text{CLIP}}(\mathbf{y})$ derived from the input prompt $\mathbf{y}$ using CLIP's text encoder. The denoising model $\epsilon_\theta(\cdot)$ is trained to predict the noise $\epsilon$ added to the latent $\mathbf{z}$ at timestep $t$ using the objective:
\begin{equation}
\mathcal{L} = \mathbb{E}_{\mathbf{z}\sim\mathcal{E}(\mathbf{T}), \mathbf{c}, \epsilon\sim\mathcal{N}(\mathbf{0},\mathbf{I}),t} \left[ \| \epsilon-\epsilon_\theta (\mathbf{z}_t, \mathbf{c}, t) \|_2^2 \right].
\end{equation}
At inference, generation starts by sampling initial noise $\mathbf{z}_T \sim \mathcal{N}(\mathbf{0},\mathbf{I})$, which undergoes iterative denoising across $T$ steps to yield $\mathbf{z}_0$, ultimately decoded into the output image $\mathcal{D}(\mathbf{z}_0)$.

The UNet employs attention mechanisms for text-image alignment through cross-attention and self-attention layers. The cross-attention layer establishes token-to-region correspondence by projecting the text embedding $\mathbf{c}$ into keys $\mathbf{K}$ and values $\mathbf{V}$, while queries $\mathbf{Q}$ originate from the UNet's intermediate features. The cross-attention distribution $\mathbf{A}^{\mathbf{c}}$ is computed as:
\begin{equation}
\mathbf{A}^{\mathbf{c}} = \text{softmax}\left( \frac{\mathbf{Q}\mathbf{K}^\top}{\sqrt{d}} \right),
\end{equation}
where $\mathbf{A}^{\mathbf{c}}_{y_i}(x,y)$ denotes the attention probability assigned to text token~$y_i$ at spatial location~$(x,y)$. Similarly, the self-attention layer~$\mathbf{A}^{s}$ captures spatial relationships within image features, with $\mathbf{A}^{s}_{x,y}$ representing the attention map for position~$(x,y)$.

\subsection{LatentMapper}
LatentMapper is proposed to initial noise optimization, which addressing semantic misalignments in text-to-image generation through latent space partitioning and distribution-aware optimization. This approach first partitions the latent space into valid and invalid regions using geometrically interpretable attention diagnostics.

Initial latent space partitioning identifies noise suitability through geometrically interpretable metrics. LatentMapper uses the cross-attention response to score $\mathcal{S}_{\text{CrossAttn}}$ to quantify subject neglect by measuring activation strength for target tokens $\mathbf{T}$ in prompt $\mathcal{Y}$:
\begin{equation}
\mathcal{S}_{\text{CrossAttn}} = 1-\min_{\mathbf{y}_i \in \mathcal{Y}} \max(\mathbf{A}^{\mathbf{c}}_{\mathbf{y}_i})
\end{equation}
Scores below threshold $\tau_c$ ensure all subjects receive sufficient activation. 

LatentMapper also uses the self-attention conflict to score $\mathcal{S}_{\text{SelfAttn}}$ measures spatial entanglement between subjects. For text token pairs $(y_i, y_j)$, spatial centroids are located via $(x_i,y_i) = \arg \max \mathbf{A}^{\mathbf{c}}_{\mathbf{y}_i}$, followed by overlap computation of their self-attention maps:
\begin{equation}
\mathbf{f}(\mathbf{y}_i,\mathbf{y}_j) = \frac{\sum_{x,y} \min(\mathbf{A}^s_{x_i,y_i}, \mathbf{A}^s_{x_j,y_j})}{\sum_{x,y} (\mathbf{A}^s_{x_i,y_i} + \mathbf{A}^s_{x_j,y_j})}, \quad 
\mathcal{S}_{\text{SelfAttn}} = \sum_{i<j} \frac{\mathbf{f}(\mathbf{y}_i,\mathbf{y}_j)}{N}
\end{equation}
where $N$ denotes the number of subject pairs. Scores below $\tau_s$ confirm distinct subject separation. Noise is valid when both $\mathcal{S}_{\text{CrossAttn}} < \tau_c$ and $\mathcal{S}_{\text{SelfAttn}} < \tau_s$ hold simultaneously.

The optimization navigates arbitrary noise toward valid regions by parameterizing a learnable Gaussian distribution $\mathcal{N}(\bm{\mu}, \bm{\Sigma})$. Noise is transformed as:
\begin{equation}
\bm{\epsilon}' = \bm{\mu} + \bm{\Sigma}^{1/2} \mathbf{z}_T, \quad \mathbf{z}_T \sim \mathcal{N}(\mathbf{0},\mathbf{I})
\end{equation}
where $\bm{\mu}$ and $\bm{\Sigma}$ are optimized via a joint loss:
\begin{equation}
\mathcal{L}_{\text{joint}} = \mathcal{S}_{\text{CrossAttn}} + \mathcal{S}_{\text{SelfAttn}} + \lambda \cdot \text{KL}\left( \mathcal{N}(\bm{\mu},\bm{\Sigma}) \parallel \mathcal{N}(\mathbf{0},\mathbf{I}) \right)
\label{eq:joint_loss}
\end{equation}
The first two terms minimize subject neglect and mixing, while the Kullback-Leibler divergence constrains the optimized distribution to remain proximate to the original Gaussian prior.

The optimization proceeds in two stages:  

\textbf{Inner optimization loop:} For each sampling round, initialize $\bm{\mu} = \mathbf{0}$, $\bm{\Sigma} = \mathbf{I}$, and $\mathbf{z}_T \sim \mathcal{N}(\mathbf{0},\mathbf{I})$. Then iteratively update parameters using the $\mathcal{L}_{\text{joint}}$ and gradient descent. Early termination occurs when validity criteria are satisfied.

\textbf{Outer sampling rounds:} If convergence fails in the inner loop, repeat the process for multiple independent rounds. Select the noise sample $\bm{\epsilon}'$ achieving the lowest combined score $\mathcal{S}_{\text{CrossAttn}} + \mathcal{S}_{\text{SelfAttn}}$ across all optimization trajectories. This two-stage approach balances computational efficiency with semantic fidelity.

\subsection{SimPrune}

To enhance computational efficiency in both denoising and noise optimization, we introduce SimPrune, a targeted token pruning strategy that operates exclusively on self-attention layers. This selective approach maintains cross-attention integrity to preserve text-image alignment. The selective is accelerating computations through two stages:

\textbf{Token Puring:} 
A minimal set of representative tokens is identified for guiding pruning. Given input tokens $\mathbf{T} \in \mathbb{R}^{N \times C}$ (token count $N$, channel dimension $C$), pairwise cosine similarity is computed:  
\begin{equation}
\text{CosSim}_{i,j} = \frac{\mathbf{T}_{i} \cdot \mathbf{T}_{j}}{\|\mathbf{T}_{i}\|\|\mathbf{T}_{j}\|}, \quad i,j \in [1,N]
\end{equation}  
Aggregated similarity scores per token are derived as $\text{SimScore}_{i} = \sum_{j=1}^N \text{CosSim}_{i,j}$. Gaussian noise $\epsilon \sim \mathcal{N}(0,\sigma^2)$ is injected to ensure diverse selection:  
\begin{equation}
\text{Noisy SimScore} = \text{SimScore} + \epsilon
\end{equation}

Next, which tokens can be pruned need to be found by using base tokens. The spatial distribution of base tokens is enforced through two steps. First, tokens are partitioned into non-overlapping $s \times s$ patches, where $s$ denotes the patch size hyperparameter. Then, for each patch, the token with the maximum Noisy SimScore is selected. The process can be formalized as:  
\begin{equation}
\mathbf{T}_{\text{base},p} = \arg \max_{\mathbf{T}_{m \in \mathcal{P}_p}} \text{Noisy SimScore}(\mathbf{T}_{m}), \quad p=1,\dots,P
\end{equation}
where $\mathcal{P}_p$ denotes tokens in the $p$-th patch, yielding $P = N/s^2$ base tokens.

Tokens most similar to base tokens are designated for pruning using precomputed cosine similarities:  
\begin{equation}
\mathbf{T}_{\text{prune}} = \underset{\mathbf{T}_i}{argtopK}  \max_{\mathbf{T}_j} \text{CosSim}(\mathbf{T}_i, \mathbf{T}_j) \quad \mathbf{T}_i \in \mathbf{T}-\mathbf{T}_{\text{base}},  \mathbf{T}_j \in \mathbf{T}_{\text{base}}
\end{equation} 
where $K$ controls the pruning ratio. This leverages spatial locality-tokens adjacent to base tokens exhibit minimal recovery error. The same $\mathbf{T}_{\text{prune}}$ is applied to both noise optimization and generation stages.

\textbf{Pruned Token Recovery:}  
The pruned tokens must be restored in the UNet but not in the LatentMapper. The unpruned token subset $\mathbf{T}_{\text{keep}} = \mathbf{T}-\mathbf{T}_{\text{prune}}$ is processed by neural layers to generate output features $f(\mathbf{T}_{\text{keep}})$. For each pruned token $\mathbf{T}_k \in \mathbf{T}_{\text{prune}}$, we first identify its most similar base token $\mathbf{T}_{b^*} = \arg \max_{\mathbf{T}_j \in \mathbf{T}_{\text{base}}} \CosSim(\mathbf{T}_k, \mathbf{T}_j)$ and then copy the output feature of $\mathbf{T}_{b^*}$ such that $f(\mathbf{T}_k) \leftarrow f(\mathbf{T}_{b^*})$. The final output combines the processed unpruned tokens and the recovered pruned tokens.

This approach achieves acceleration by reducing token count in computationally intensive layers, while spatial uniformity of base tokens and injected randomness prevent degradation in generation fidelity.

In LatentMapper, token pruning accelerates attention diagnostics. The self-attention conflict score $\mathcal{A}_{\text{s}}$ computation leverages the preserved tokens' spatial distribution, where overlapping regions exhibit lower similarity to other regions, thus avoiding pruning. This enables efficient assessment of spatial entanglement. Hence, attention validity checks require only relative spatial relationships.

\section{Experiment and Analysis}\label{sec:exp}  

\begin{table}[ht]
\centering
\vspace{-0.5cm}
\caption{Average CLIP Similarity (\%) for Animal-Animal Category}
\vspace{-0.3cm}
\label{tab:clip_animal_animal}
\begin{tabular}{|l|c|c|c|}
\hline
Method & Full Prompt & Minimum Object & Text-Text Similarity \\
\hline
Stable Diffusion~\cite{rombach2022high} & 31.2 & 21.6 & 76.6 \\
A-STAR~\cite{agarwal2023star} & 33.0 & 25.0 & 83.5 \\
Divide-and-Bind~\cite{li2023divide} & 33.2 & 24.5 & 82.5 \\
Attend-and-Excite~\cite{chefer2023attend} & 33.2 & 25.1 & 82.0 \\
Structure Diffusion~\cite{feng2022training} & 31.0 & 22.0 & 76.0 \\
Composable Diffusion~\cite{liu2022compositional} & 29.5 & 23.5 & 68.0 \\
Initno~\cite{guo2024initno} & 33.4 & 25.9 & 84.8 \\
Ours & \textbf{33.5} & \textbf{26.0} & \textbf{84.9} \\
\hline
\end{tabular}
\vspace{-0.8cm}
\end{table}

\begin{table}[ht]
\centering
\vspace{-0.5cm}
\caption{Average CLIP Similarity (\%) for Animal-Object Category}
\vspace{-0.3cm}
\label{tab:clip_animal_object}
\begin{tabular}{|l|c|c|c|}
\hline
Method & Full Prompt & Minimum Object & Text-Text Similarity \\
\hline
Stable Diffusion~\cite{rombach2022high} & 31.5 & 22.5 & 77.0 \\
A-STAR~\cite{agarwal2023star} & 35.5 & 26.0 & 82.5 \\
Divide-and-Bind~\cite{li2023divide} & 35.2 & 25.5 & 81.5 \\
Attend-and-Excite~\cite{chefer2023attend} & 35.8 & 26.2 & 82.0 \\
Structure Diffusio~\cite{feng2022training} & 34.0 & 24.0 & 77.0 \\
Composable Diffusion~\cite{liu2022compositional} & 33.8 & 25.0 & 76.0 \\
Initno~\cite{guo2024initno} & 36.1 & 26.4 & 83.1 \\
Ours & \textbf{36.2} & \textbf{26.5} & \textbf{83.8} \\
\hline
\end{tabular}
\vspace{-0.8cm}
\end{table}

\begin{table}[ht]
\centering
\vspace{-0.5cm}
\caption{Average CLIP Similarity (\%) for Object-Object Category}

\vspace{-0.3cm}
\label{tab:clip_object_object}
\begin{tabular}{|l|c|c|c|}
\hline
Method & Full Prompt & Minimum Object & Text-Text Similarity \\

\hline
Stable Diffusion~\cite{rombach2022high} & 33.5 & 23.8 & 75.8 \\
A-STAR~\cite{agarwal2023star} & 35.5 & 26.5 & 81.5 \\
Divide-and-Bind~\cite{li2023divide} & 35.2 & 26.2 & 80.5 \\
Attend-and-Excite~\cite{chefer2023attend} & 36.2 & 26.8 & 82.0 \\
Structure Diffusion~\cite{feng2022training} & 33.0 & 23.0 & 75.0 \\
Composable Diffusion~\cite{liu2022compositional} & 34.5 & 26.5 & 76.0 \\
Initno~\cite{guo2024initno} & 36.3 & 27.8 & \textbf{83.9} \\
Ours & \textbf{36.4} & \textbf{27.9} & \textbf{83.9} \\
\hline
\end{tabular}
\vspace{-0.8cm}
\end{table}

\subsection{Experimental settings}
We conduct experiments using the Stable Diffusion v1.5 (SD1.5) model as the baseline, which belongs to the latent diffusion model (LDM) family. The experiments are performed on NVIDIA 3090 GPUs with 24GB memory, ensuring efficient computation for both training and inference.

For implementation details, we follow the standard configuration of SD1.5 and set the guidance parameter to 7.5 for all experiments. To preprocess attention maps, a Gaussian filter with a 3×3 kernel and standard deviation of 0.5 is applied to smooth the cross-attention and self-attention maps. The denoising process is configured with a total timesteps \( T = 50 \), following the protocol of DDPM-based inference.

We evaluate our method on the Animal-Animal,
Animal-Object, and Object-Object datasets~\cite{chefer2023attend}, which contain prompts with two subjects and color attributes. Additionally, our approach is validated on complex prompts with multiple subjects and attributes, as discussed in subsequent sections.

\subsection{Qualitative comparison}
We compare the visual quality of images generated by SD1.5~\cite{rombach2022high}, Composable Diffusion~\cite{liu2022compositional}, Structure Diffusion~\cite{feng2022training}, Attend-and-Excite~\cite{chefer2023attend}, Divide-and-Bind~\cite{li2023divide}, A-STAR~\cite{agarwal2023star} and InitNO~\cite{guo2024initno}, and our method under identical text prompts and same seeds. For fair comparison, our method and SiTo are integrated with our token pruning module under the identical pruning ratio, a hyperparameter~$\gamma=0.4$, where our method applies this module during both noise optimization and generation stages, while SiTo employs it only during the generation stage.

Our approach demonstrates notably stronger generalization capabilities across diverse prompting scenarios, including complex scenes with multiple interacting objects. By maintaining consistent semantic alignment and structural coherence regardless of prompt complexity, it reliably produces high-fidelity outputs that adhere to the input description’s intent.

\begin{figure}[htbp]
    \centering
    \includegraphics[width=1.0\textwidth]{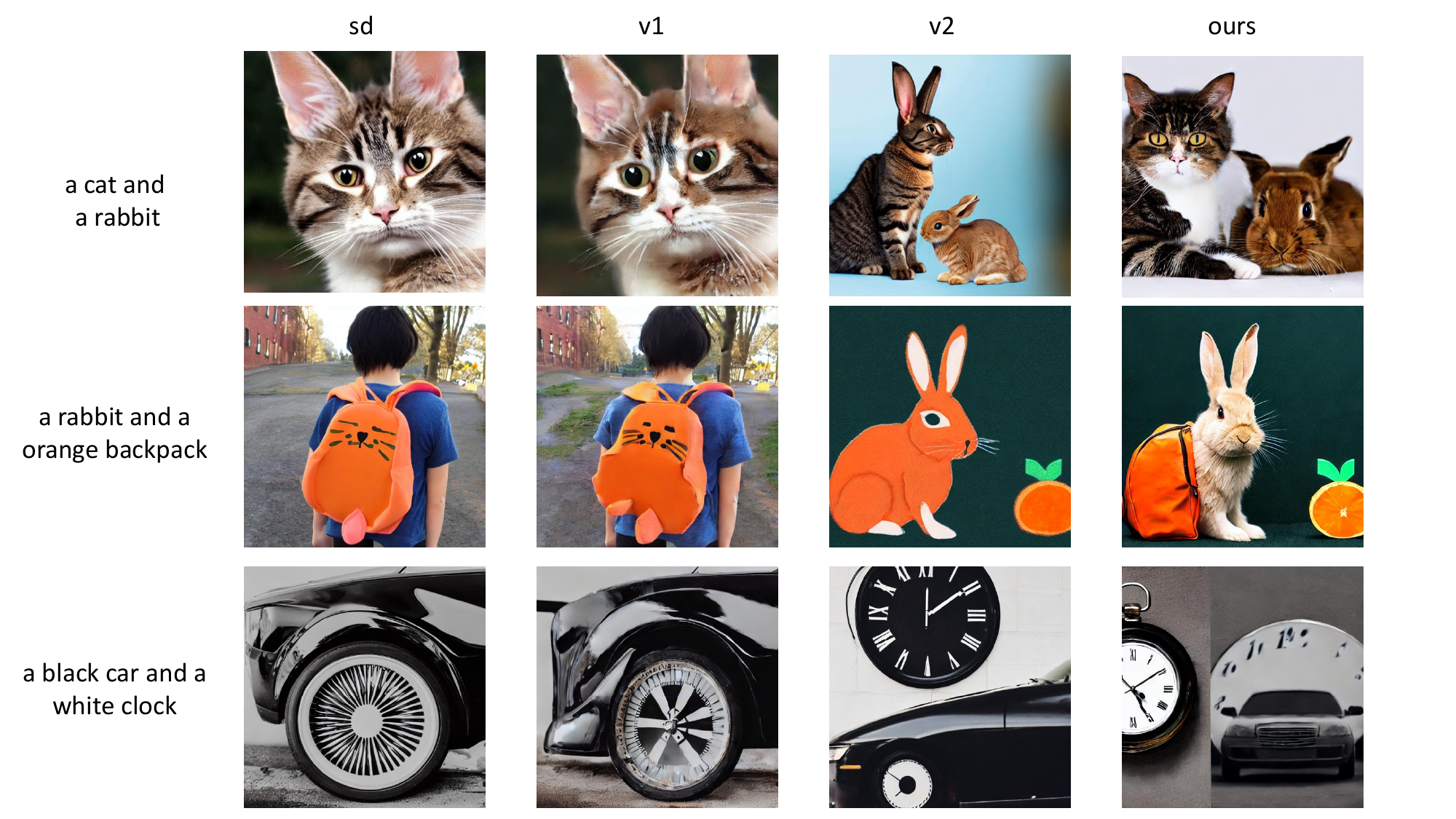}
    \caption{Visualization of ablation study.For all methodologies, images are synthesized using same text prompts and the same seed. Our approach demonstrates strong correspondence with text prompts, all the while preserving high visual realism in generated outputs.}
    \label{fig2}
\end{figure}

\begin{figure}[t]
    \centering
    \includegraphics[width=1.0\textwidth]{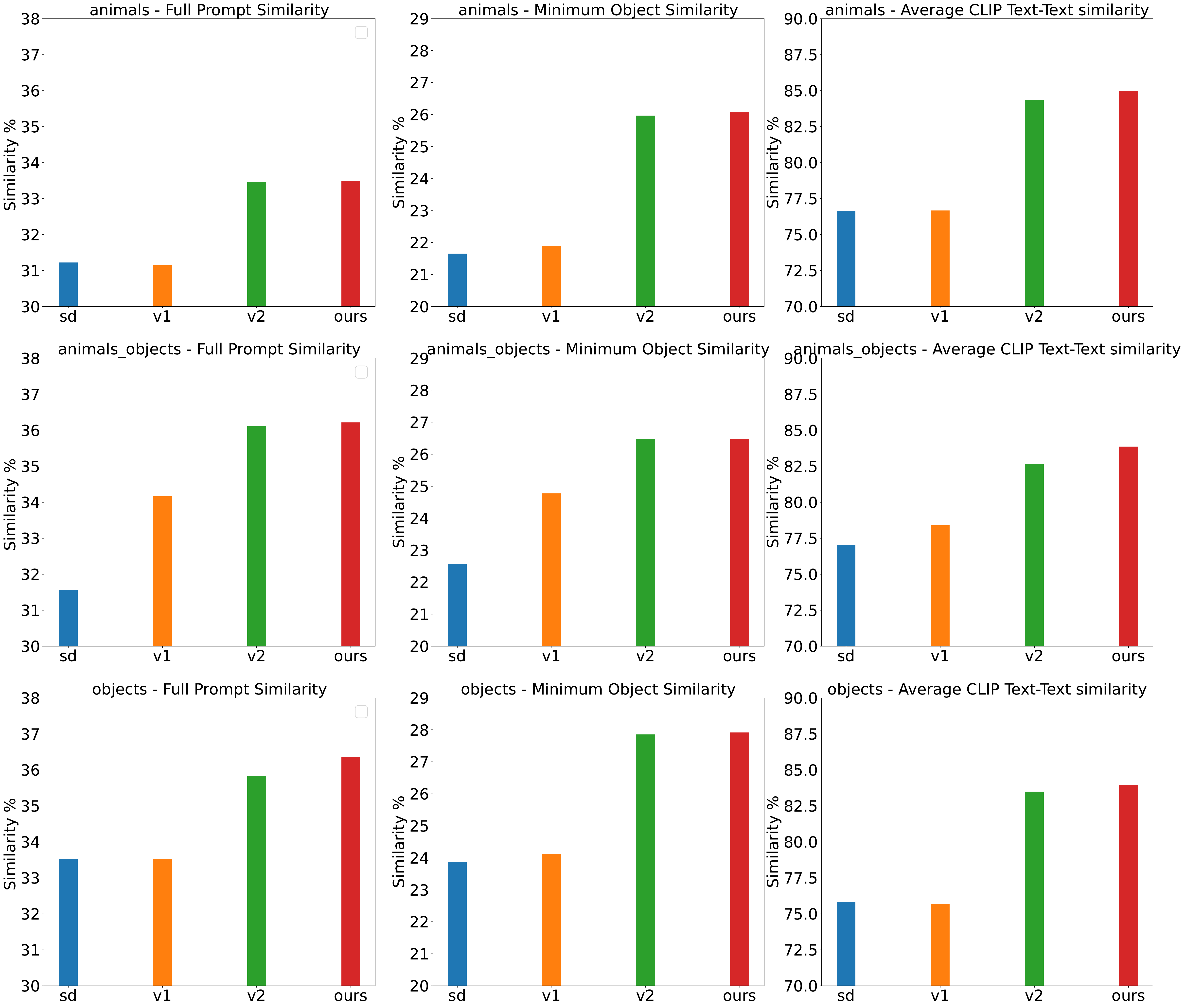}
    \caption{Average CLIP Image-Text Similarity, including Full Prompt Similarity and Minimum Object Similarity, and Average CLIP Text-Text Similarity are reported for the quantitative measurement. Higher is better.} \label{fig3}
\end{figure}

\subsection{Quantitative Evaluation}

We conduct quantitative evaluations of the proposed method using CLIP's image-text similarity and text-text similarity metrics. For each prompt, 16 images are generated, and the average CLIP cosine similarity is calculated while maintaining consistent random seeds across all compared methods. Specifically, for CLIP image-text similarity, we report the full-prompt similarity (cosine similarity between the complete prompt and the generated image) and the minimum object similarity (the smallest similarity value between the image and each of the two subject prompts). For CLIP text-text similarity, we generate captions for the images using BLIP~\cite{li2022blip} and compute the similarity between these captions and the input prompt. The quantitative results in Tables~\ref{tab:clip_animal_animal}, \ref{tab:clip_animal_object}, and \ref{tab:clip_object_object} comprehensively demonstrate the effectiveness of our approach across diverse prompt categories.

For Animal-Animal (Table~\ref{tab:clip_animal_animal}), our method achieves the highest scores in Full Prompt Similarity (33.5\%), Minimum Object Similarity (26.0\%), and Text-Text Similarity (84.9\%), outperforming InitNO by 0.1\% in both Full Prompt and Minimum Object metrics, which shows that our method's excellent ability to handle inter object interactions. For example, when generating an image of "a cat and a lion", our approach can clearly distinguish between the two animals. This enhancement is due to the fact that our LatentMapper can obtain Valid Initial Noise, and our SimPrune can ignore unimportant parts during the image generation process.

For Animal-Object (Table~\ref{tab:clip_animal_object}), our approach attains 36.2\% (Full Prompt), 26.5\% (Minimum Object), and 83.8\% (Text-Text), exceeding all baselines. Notably, it achieves a 0.1\% gain over InitNO in Minimum Object Similarity and a 0.7\% lead in Text-Text Similarity, reflecting the model's improved text-semantic alignment. For instance, when generating an image of "an elephant with a crown", our method can accurately place the crown on the elephant's head. The crown is clearly associated with the elephant, and there is no mis-attribution or blurring of the relationship between the two.

In Object-Object (Table~\ref{tab:clip_object_object}), our method attains 36.4\% (Full Prompt) and 27.9\% (Minimum Object), surpassing InitNO by 0.1\% in both metrics. In cases like "a purple balloon and a white clock", our approach can ensure that the purple balloon and the white clock are clearly separated in space. Each object has its own distinct boundaries, and there is no overlapping or merging of the two objects. The purple color of the balloon does not bleed onto the white clock, and the details of both objects are well-preserved, presenting a clear and accurate image.

\subsection{Quantitative Evaluation} 

We conduct quantitative evaluations to assess the semantic alignment performance of our method. We adopt CLIP-based similarity metrics, including image-text similarity and text-text similarity, which are commonly used to measure prompt-image consistency in text-to-image generation tasks.
For each input prompt, we generate 16 images using our method and all baselines, with identical random seeds to ensure fair comparisons. We compute the average cosine similarity over all samples for each metric.
For image-text similarity, we report two metrics. The full-prompt similarity measures the cosine similarity between the entire prompt and the generated image using CLIP encoders, reflecting overall semantic alignment. The minimum object similarity computes the lowest similarity between the image and each subject entity in the prompt, capturing object-level consistency and penalizing missing or inaccurate objects.
For text-text similarity, we generate captions for each image using BLIP and compute their cosine similarity with the original prompt using CLIP’s text encoder. This evaluates whether the image content, as interpreted in language, matches the input description.
These metrics collectively provide a comprehensive assessment of both visual and linguistic alignment between prompts and generated images.

\subsection{Ablation study} 
We investigate several variants to prove the importance of our propose modules in OptiPrune. (1) V1 removes the LatentMapper. (2) V2 removes the SimPrune.Quantitative and qualitative comparisons against the full OptiPrune reveal critical insights into each component's functionality.

Visual results under identical prompts and random seeds in Figure \ref{fig2} clearly demonstrate the distinct roles of the ablated modules. Removing the LatentMapper (V1) leads to severe subject mixing: in "a cat and a rabbit", it generates hybrid creatures with blended feline/lapine features; in "a rabbit and an orange backpack", it produces rabbit-featured backpacks; and in "a black car and a white clock", it fuses the clock with car wheels. These results confirm LatentMapper's critical role in resolving semantic conflicts and maintaining distinct subject identities. Conversely, ablating SimPrune (V2) causes noticeable distortion and incorrect attribute binding. This manifests as: subtle rabbit features appearing on cats in "a cat and a rabbit"; orange coloration erroneously applied to rabbits instead of backpacks in "a rabbit and an orange backpack"; and misassigned colors (e.g., black clocks) in "a black car and a white clock". V2 further compromises spatial separation between subjects, highlighting SimPrune's essential role in precise attribute binding and object disentanglement.

Figure \ref{fig3} quantitatively validates these observations through comprehensive CLIP-based metrics. The superior Full Prompt Similarity of our full model directly corresponds to its ability to avoid hybrid creatures and color misassignments , ensuring all attributes align with textual descriptions. Notably, the Minimum Object Similarity metric specifically captures subject mixing (e.g., cat-rabbit hybrids) and spatial entanglement (e.g., clock-wheel fusion). Furthermore, the Text-Text Similarity gap highlights our modules' synergistic role in disentangling conflicting semantics: LatentMapper prevents cross-category feature bleeding (animal-object boundaries), while SimPrune resolves intra-prompt attribute competition. Across all metrics and scenarios, “ours” consistently attains the highest similarity scores, corroborating the synergistic efficacy of the LatentMapper and SimPrune modules.

\section{Conclusion}\label{sec:con} 
This paper proposed OptiPrune, a unified framework integrating distribution-aware initial noise optimization with hardware-friendly similarity-based token pruning to enhance semantic alignment and computational efficiency in text-to-image diffusion models. Experiments on benchmarks validated OptiPrune’s superiority in prompt-image consistency and latency reduction, outperforming methods like InitNO and SiTo in CLIP similarity metrics. In terms of limitations, although token pruning accelerates the process, the initial noise optimization still incurs non-negligible computational overhead. For future directions, we will focus on exploring adaptive pruning strategies specifically tailored for initial noise optimization and investigating multimodal alignment techniques to extend our framework to video diffusion models.

\end{document}